\documentclass[sigconf, 9pt]{acmart}
\usepackage{graphicx}
\usepackage{hyperref} 
\usepackage{wrapfig}
\usepackage{pifont}
\usepackage{tikz}
\usepackage{comment}
\usepackage{color}
\usepackage{booktabs} 
\usepackage{multirow,mathtools} 
\usepackage{algorithm,algpseudocode}
\usepackage{threeparttable}
\usepackage{color, colortbl}

\newcommand{\NA}{---}
\definecolor{Gray}{gray}{0.9}

\AtBeginDocument{%
  }

\copyrightyear{2023}
\acmYear{2023}
\setcopyright{rightsretained}
\acmConference[ASPDAC '23]{28th Asia and South Pacific Design Automation Conference}{January 16--19, 2023}{Tokyo, Japan} \acmBooktitle{28th Asia and South Pacific Design Automation Conference (ASPDAC '23), January 16--19, 2023, Tokyo, Japan}\acmDOI{10.1145/3566097.3567864} \acmISBN{978-1-4503-9783-4/23/01}

\begin{document}

\title{Data-Model-Circuit Tri-Design for Ultra-Light Video Intelligence on Edge Devices}

\author{Yimeng Zhang}
\authornote{Authors contributed equally to this research.}
\affiliation{%
  \institution{Michigan State University, USA}}

\author{Akshay Karkal Kamath}
\authornotemark[1]
\affiliation{\institution{Georgia Institute of Technology, USA}}

\author{Qiucheng Wu}
\authornotemark[1]
\affiliation{\institution{UC, Santa Barbara, USA}}

\author{Zhiwen Fan}
\authornotemark[1]
\affiliation{\institution{University of Texas at Austin, USA}}

\author{Wuyang Chen}
\affiliation{\institution{University of Texas at Austin, USA}}

\author{Zhangyang Wang}
\affiliation{\institution{University of Texas at Austin, USA}}

\author{Shiyu Chang}
\affiliation{\institution{UC, Santa Barbara, USA}}

\author{Sijia Liu}
\affiliation{\institution{Michigan State University, USA}}

\author{Cong Hao}
\affiliation{\institution{Georgia Institute of Technology, USA}}

\renewcommand{\shortauthors}{Zhang et al.}

\settopmatter{printacmref=true} 

\begin{abstract}
  In this paper, we propose a \textit{data-model-hardware tri-design} framework for high-throughput, low-cost, and high-accuracy multi-object tracking (MOT) on High-Definition (HD) video stream.  First, to enable ultra-light video intelligence, we propose temporal frame-filtering and spatial saliency-focusing approaches to reduce the  complexity  of massive video data. Second, we exploit structure-aware weight sparsity to design a hardware-friendly model compression method. Third, assisted with data and model complexity reduction, we propose a sparsity-aware, scalable, and low-power accelerator design, aiming to deliver real-time performance with high energy efficiency. Different from existing works, we make a solid step towards the synergized software/hardware co-optimization for realistic MOT model implementation. 
  Compared to the state-of-the-art MOT baseline, our tri-design approach can achieve {12.5$\times$} latency reduction, 20.9$\times$ effective frame rate improvement, $5.83\times$ lower power, and $9.78\times$ better energy efficiency, without much accuracy drop. 
\end{abstract}

\maketitle

\section{introduction}
The past decade has  witnessed a tremendous success of deep neural networks (DNNs) 
\cite{alom2018history}.
In this paper, we focus on 
DNN-based multi-object  tracking  (MOT), which has been used as a technological basis for  video intelligence  and is drastically improving the quality of human life, \textit{e.g.},   autonomous driving (AD)   \cite{yurtsever2020survey}.
Despite a proliferation of MOT techniques \cite{kalake2021analysis}, little research effort has been made towards   the \textit{data-efficient} and \textit{co-design solutions} for MOT across the full software/hardware stack. Our work   aims to close this gap by
developing a high-throughput, low-cost, and high-accuracy video processing  algorithm/hardware pipeline. 
We provide a holistic viewpoint on how contemporary MOT methods  can be renovated   to  deal with massive quantities of data and high-complexity DNNs.

Despite extensive existing works, data and model efficiencies of MOT are mainly  studied either from   the algorithm perspective or  from the model design perspective. For example, in \cite{wang2020tracking}, the MOT efficiency is improved by leveraging meta learning \cite{finn2017model} to solve an instance detection problem.
In addition to algorithms,  several works have focused on developing model compression techniques (such as weight pruning and weight quantization) to  reduce the model complexity of MOT \cite{stacker2021deployment,muhawenayo2021compressed}. However, the efficiency  of ``compressed'' models are evaluated  without considering the practical hardware platform, such as low-power FPGAs. 
Known FPGA acceleration techniques ~\cite{hao2019codesign,zhang2018dnnbuilder}
do not explore data sparsity or model sparsity, thereby missing huge optimization opportunities. Additionally, existing accelerators are evaluated on the ImageNet dataset with small input image sizes and do not scale to real-world High-Definition (HD) video frames, limited by the device and/or design tool capabilities. To the best of our knowledge, there is no prior work that discusses efficient implementation of MOT on the edge for HD video processing by fully utilizing data- and model-level sparsity.

\noindent \textbf{Contributions.}
To close the   gap between MOT algorithm design and efficient implementation,
we propose a \textit{data-model-hardware tri-design} approach.
We summarize our contributions as follows.

\noindent $\bullet$ \textbf{Dynamic frame filtering}. We propose a reinforcement learning-based lightweight algorithm  to achieve temporal data reduction. 

\noindent  $\bullet$ \textbf{Spatial attention focusing}. We propose a saliency-guided spatial data reduction method to  eliminate uninformative pixels from both the input frames and the intermediate feature maps. 

\noindent  $\bullet$ \textbf{Hardware-aware model compression}. We leverage kernel-wise pattern-aware sparsity of an MOT model to achieve hardware-friendly model compression.  

\noindent $\bullet$ \textbf{Tri-design implementation and evaluation}. 
We implement the proposed tri-design framework on a hardware platform comprising of one Xilinx ZCU104 and two Xilinx Alveo U50 FPGAs and conduct extensive experiments to evaluate its effectiveness in both accuracy and on-device efficiency metrics.
We demonstrate that  our approach can achieve 
{12.5$\times$}
latency reduction, 20.9$\times$ effective frame rate improvement, $5.83\times$ lower power, and $9.78\times$ better energy efficiency, 
without much accuracy drop compared to the state-of-the-art MOT baseline \cite{pang2021quasi} on the BDD100K dataset \cite{yu2020bdd100k}.

\section{Preliminaries and Motivation}
\label{sec:motivation-hw}

In this paper, we focus on MOT (multiple-object tracking) which deals with massive video data. 
Current MOT methods are commonly built upon DNNs 
and follow the paradigm of tracking-by-detection 
\cite{wang2020towards}.
QDTrack~\cite{pang2021quasi} is the state-of-the-art MOT algorithm introduced on the BDD100K dataset~\cite{yu2020bdd100k}, one of the most representative and challenging self-driving car datasets.
QDTrack employs Faster R-CNN~\cite{ren2015faster} with Feature Pyramid Network (FPN)~\cite{lin2017feature} as an object detection backbone network, leverages Contrastive Learning \cite{wu2018unsupervised} to optimize the backbone network parameters, and utilizes Bi-directional Softmax \cite{pang2021quasi} in the embedding space for object association and tracking. 
The focus of this work is \textit{not} to develop new MOT model algorithms but to instead build a high-throughput, low-cost, and high-accuracy video processing \textit{implementation pipeline} across the \textit{full data-processing/DNN operation/hardware stack.} 

The computational landscape of  MOT has been rapidly evolving on dedicated computing hardware (\textit{e.g.}, FPGAs and ASICs)
~\cite{zhang2021yoloFPGA}.
Despite the advancements of hardware accelerators~\cite{li2019implementing}
 and software/hardware co-design techniques~\cite{hao2019codesign},
 \textbf{critical limitations} still exist for real-time video processing. \underline{First}, there is a sharp increase not only in ML model size but also computation complexity, especially for HD video frames. 
With HD inputs, any two $3\times3$ convolution layers of ResNet-50 already have higher complexity than the entire ResNet-50 with ImageNet inputs. This asserts huge pressure for real-time inference on hardware. \underline{Second}, both the hardware devices and design tools exhibit poor scalability. For example, the most powerful FPGA till date, Xilinx Alveo U55C, has 9,024 DSP slices. It can provide peak performance of 2,256 GOPs for QDTrack (157.9 GOPs), but the performance upper bound is 70 ms per frame, which is far from real-time requirement. Meanwhile, the design tools are strictly limited by the largest parallelism that can be supported. For example, Xilinx Vitis HLS tool constrains that the parallelism factor should be lower than 4096~\cite{aarrestad2021fast}, while Intel Design Toolkit does not support more than 256 concurrent
kernels.
With larger parallelism, hardware designs can easily take days or even weeks to synthesize, which drastically increases the development cycle and the time-to-market (TTM).
Consequently, there is still a huge gap between the low-power real-time video processing requirements and the limited capability of existing hardware acceleration techniques.

As inspired by above, our work advances the design of practical MOT systems by tackling the following \textbf{challenges (C1--C3)}. \textbf{(C1)}:  The massive video data and their high frame rates make the implementation of MOT prohibitively expensive in energy and latency. Thus, our first goal is to develop novel data complexity reduction techniques that can fully exploit the  temporal and saliency redundancy of video frames over time and   space.  
\textbf{(C2)}:
SOTA DNN-based MOT models typically contain gigantic number of weights making them poor candidates for hardware implementation. To this end, we propose a hardware-friendly model compression technique for MOT. 
\textbf{(C3)}:
A holistic data-model-hardware full-stack design approach is lacking. It remains elusive how current MOT solutions can be applied subject to Small Size, Weight, and Power (SWaP) constraints in practice. 
To the best of our knowledge, there is no prior work that addresses the challenges (C1--C3) in a single unified MOT implementation pipeline.  

\section{Methodology 
}

\newcommand*\circled[1]{\raisebox{.4pt}
                    {\tikz[baseline=(char.base)]{
            \node[shape=circle,draw,inner sep=1pt, style={fill=black, text=white}, scale=0.75] (char) {\textbf{#1}};}}}

\begin{figure}[t]
\vspace*{-1mm}
    \centering
    \includegraphics[width=1\linewidth]{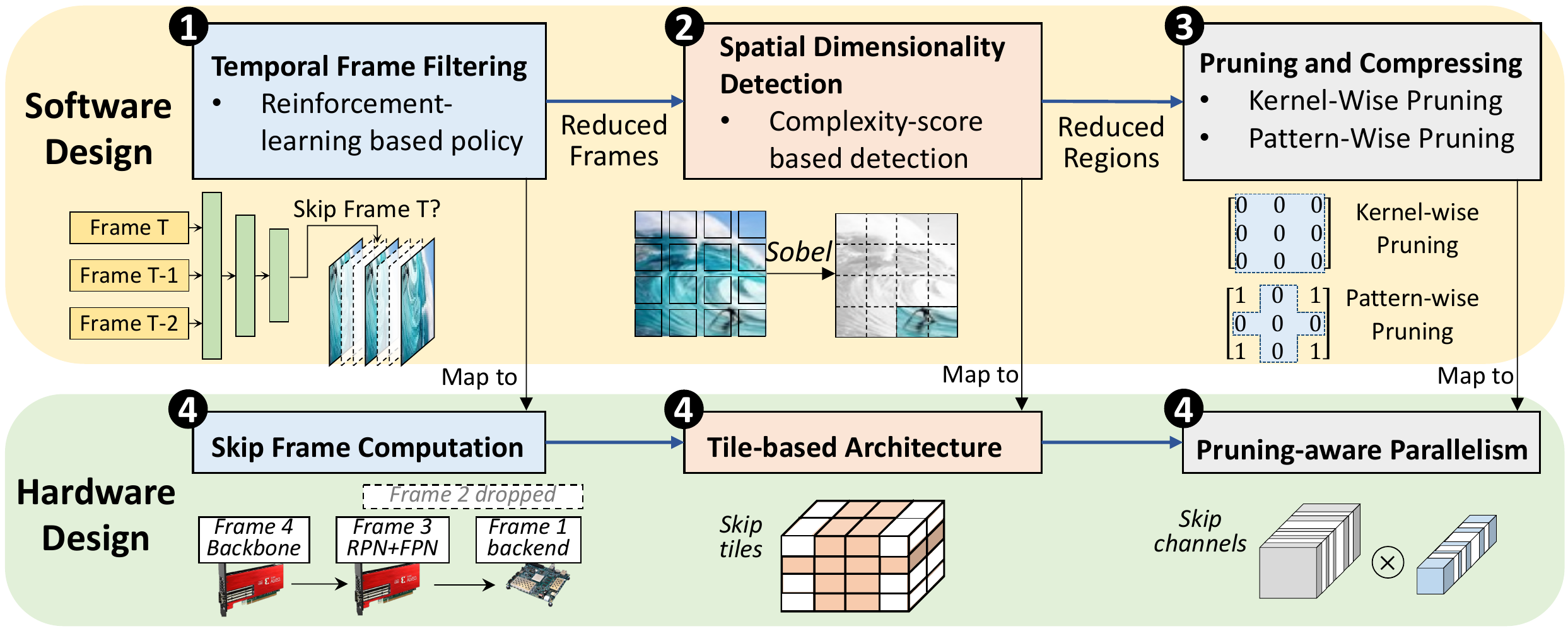}
    \vspace*{-4mm}
    \caption{
    Overview of data-model-hardware tri-design.
    } 
\vspace*{-5mm}
     \label{fig:framework}
\end{figure}
            
\subsection{{Overview of proposed approach.}}
The proposed data-model-hardware tri-design   includes four main components as shown in Fig. \ref{fig:framework}. \textbf{\circled{1} (Data) Temporal frame filtering}. It is achieved using  reinforcement   learning (RL) to filter unnecessary frames at the front-end data processing stage and eventually   reduce processing power along with latency costs on hardware. \textbf{\circled{2} (Data) Spatial saliency focusing}. It   
provides the second-stage data complexity reduction by peering into   the salient region of a frame  to eliminate uninformative pixels. 
\textbf{\circled{3} (Model) Hardware-friendly deep model pruning}. It exploits the structured sparsity patterns of weights to reduce the model size on hardware without much loss of accuracy.  
\textbf{\circled{4} (Hardware) Backend acceleration}. Guided by the algorithmic pipeline, it adopts low-power FPGAs to  achieve realistic implementation of MOT with high throughout, low latency, and high energy efficiency. 
In a nutshell, components \circled{1} and \circled{2} are designed for tackling the  data complexity challenge (in response to C1); Component \circled{3} reduces the complexity of back-end model (in response to C2); Component \circled{4}  unveils the practical efficiency of hardware solutions for MOT (in response to C3). 
 
 \subsection{{\underline{Data} complexity reduction}}

\subsubsection{RL-based temporal frame filtering}
The embedded MOT solution requires a high throughput in terms of the number of frames that can be processed in unit time. The frame processing rate should also consider the practical latency cost. Considering the massive quantities of 
video frames, achieving a video processing rate of less than 30ms (for real-time inference) becomes  implausible even on high-end FPGAs.  Due to this hardware  constraint, it is necessary to reduce the data complexity through the lens of temporal video frame dropping. Particularly,
in the AD (autonomous driving) datasets, consecutive video frames may contain   largely-overlapped  frame content such as street and road scenes. Thus, temporal frame filter is not only necessary but also reasonable for real-time MOT. 
In our preliminary experiments,  we  test a   random filtering policy with different dropping ratios. As shown in Fig.\,{\ref{fig: compare_combine}}, 
dropping frames randomly significantly hampers the object tracking performance. 
The incapability of random filtering drives us to  design a lightweight learning-based solution that can self-adjust the frame dropping scheme based on the task-end tracking performance. 

Inspired by above, {we propose} an RL-based policy network for temporal frame filtering. 
 Our policy network takes the current frame and the difference between the current frame and the previous frame as inputs. At the beginning of the policy network, we adopt a lightweight 3-layer convolutional network to extract a 2D feature map with the same height and width as the input frames. 
 In compensation with a shallow network, we adopt 5$\times$5 kernels at each convolution layer. The numbers of filters of the three convolution layers are 16, 8, and 1. Finally, with a pooling layer and sigmoid layer, we obtain the probability 
 of dropping the current frame, which indicates the salience of the current input frame.

We optimize this policy model through a regularized policy gradient method. Specifically, given a video sequence $\mathbf v$ with $n$ frames $\{v_0, v_1, ..., v_n\}$, the  policy model $G$ generates a frame importance score $G(v_i) \in (0,1)$ for each frame $v_i$.  $G(v_i)$ serves as the probability to retain the current frame and determines  the new video sequence $\mathbf v^\prime$ with $n^\prime$ frames that only consists of the selected video frames.  
The training objective function of $G$ is given by

\vspace*{-3mm}
{\small \begin{equation}
    J = -R(\mathbf v^\prime)\sum_{i=1}^n\log p(G(v_i))+ \alpha \sum_{i=1}^n (G(v_i)-\mu)^2,
    \label{eq: obj}
\end{equation}}
\vspace*{-2mm}

\noindent where $R(\mathbf v^\prime)$ is the reward of the updated video sequence $\mathbf v^\prime$ and
$p(G(v_i))$ is the probability to sample an individual frame $v_i$. We use the number of ID Switches  per frame (\textit{i.e.}, {IDSw} in Sec.\,\ref{sec: setup_exp})   as the reward function to evaluate the quality of frame dropping:
$
 R(\mathbf v^\prime) = -\mathrm{IDSw}(\mathbf v^\prime)/n^\prime 
$,
where   $n^\prime$ is the number of remaining frames in the video sequence $\mathbf v^\prime$. In \eqref{eq: obj}, inspired by  \cite{zhou2018deep}, we also regularize the salience scores $G(v_i)$ with a hyperparameter $\mu$ to avoid the case of no frame dropping  and obtaining high reward for free.

\subsubsection{Spatial saliency focusing.}

Based on the first-stage  temporal filtering, we next propose the second-stage data reduction, which \textit{attentively focuses} on the salient region of a frame and its feature map to eliminate uninformative pixels.

Towards the goal of spatial reduction, our MOT model (\textit{i.e.}, QDTrack) achieves spatial saliency by detecting and characterizing important spatial signals while monitoring high-rate video streams with low latency.  The output is a localized sub-image for each frame, on which the subsequent backend algorithm on FPGA can focus.
We propose a \textit{spatial saliency reduction} (SSR) method. First, it decomposes an input image 
into multiple patches,
and then adopts the Sobel operator
~\cite{jin2009edge} 
to compute an approximate saliency score of each  patch. This   operator exploits two 3 $\times$ 3 convolution kernels on the given patch to calculate the saliency values (\textit{i.e.}, derivatives). 
Once  the patch-wise saliency score is obtained, we smoothen it using its four neighboring patches and create a binary mask indicating the patches of an input frame that should be discarded. 

\begin{figure*}
\centering
\begin{tabular}{c c c c}
    \includegraphics[width=0.22\textwidth]{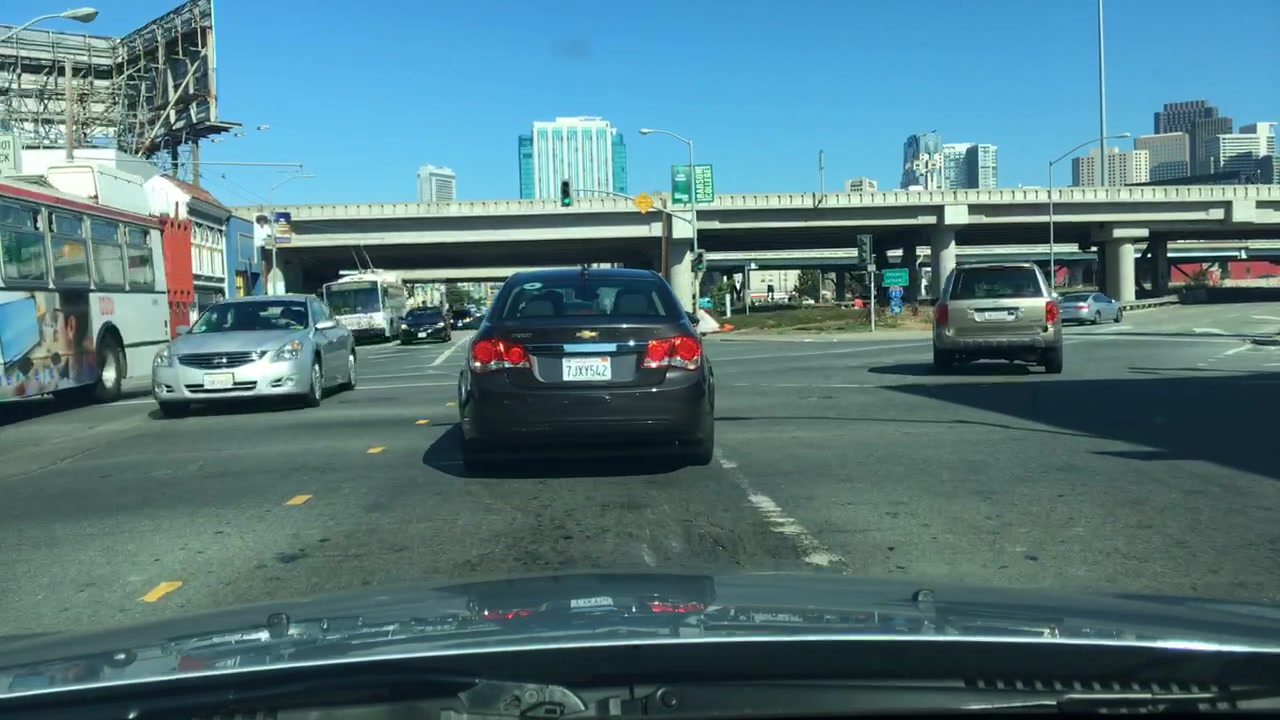} &
   \includegraphics[width=0.22\textwidth]{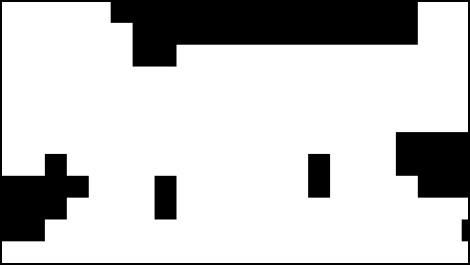} &
   \includegraphics[width=0.22\textwidth]{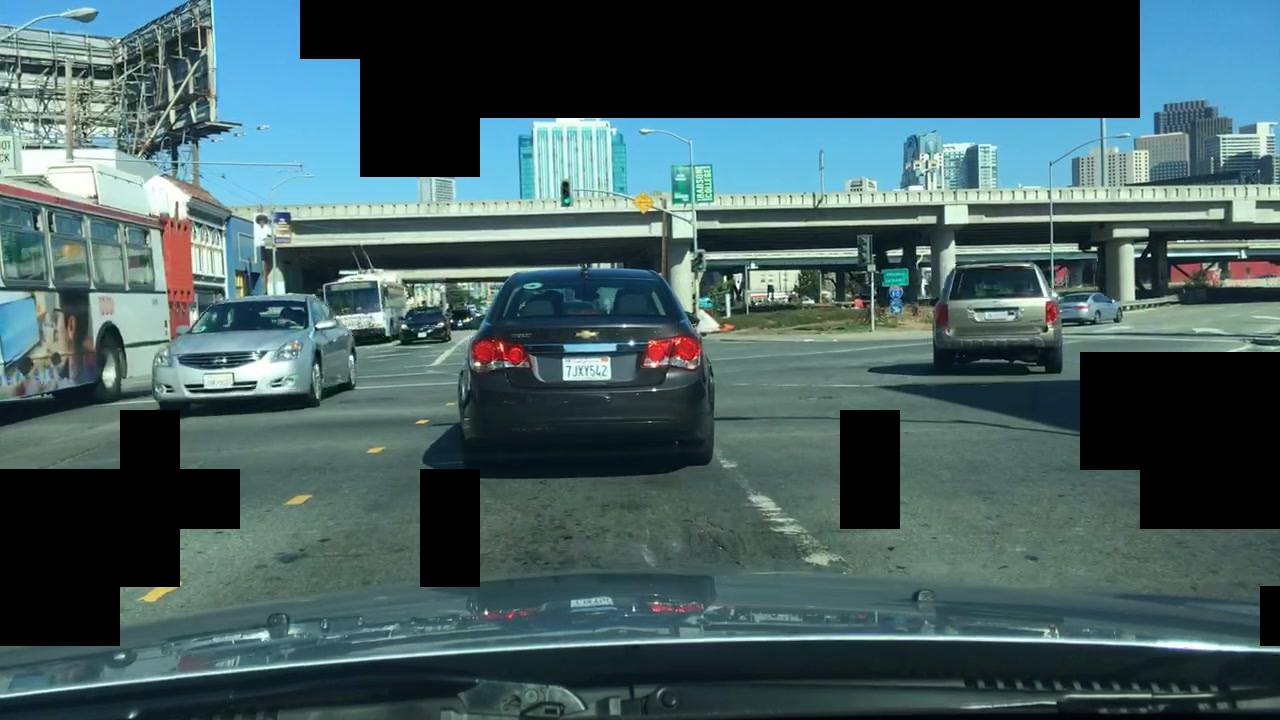} &
   \includegraphics[width=0.22\textwidth]{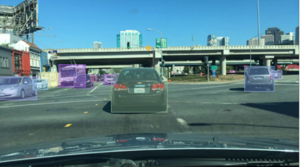} \\
   (a) Input frame & (b) Saliency Mask & (c) Masked Input & (d) GT 
\\
   \includegraphics[width=0.22\textwidth]{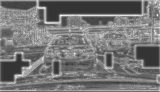} &
   \includegraphics[width=0.22\textwidth]{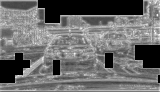} & \includegraphics[width=0.22\textwidth]{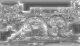} &
   \includegraphics[width=0.22\textwidth]{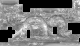} 
\\
     (e) FM(2,0)   & (f) FM(2,0) w/ Mask &(g) FM(3,0)   &
     (h) FM(3,0) w/ Mask \\
\end{tabular}
\vspace{-3mm}
\caption{Visualizations of input frame patch dropping (first row) and feature map patch dropping (second row). 
(a): An input frame with Ground Truth (GT) object bounding box shown in (d).
(b)-(c): An input saliency mask and the masked input frame, where the patch dropping ratio is set to 20\%, and the discarded pixels are marked in black. 
(e)-(f): Feature  maps in layer 0 of block 2 within backbone (ResNet-50), denoted by FM(2,0), without and with feature patch dropping mask, respectively. 
(g)-(h): Feature  maps in layer 0 of block 3, denoted by FM(3,0), without and with feature patch dropping mask, respectively. 
We can observe that the proposed spatial data reduction approach could accurately identify the redundant patch regions associated with background scenes, \textit{e.g.}, sky and street. 
}\label{fig:patch2feature}
\vspace{-5mm}
\end{figure*}

In addition to input saliency dropping, 
we further propose \textit{feature saliency dropping}, where patches of intermediate features, i.e., activations, can also be filtered. 
Our motivations are two-fold. \underline{First}, even if an input patch has
been dropped, its corresponding feature region might still have responses “leaked” from neighborhood
unfiltered areas, caused by filter stride, padding, etc.
\underline{Second}, some image regions might  be recognized as  redundant  at the intermediate feature level rather than  at the beginning input level.  As shown in Fig.\,\ref{fig:patch2feature}(e) and (g), even if a region is not dropped initially, it leads to low responses at later stages. Therefore, reducing spatial redundancy beyond the input level to feature level (Fig.\,\ref{fig:patch2feature}(f) and (h)) is valuable and is also beneficial to hardware implementation.
Feature map region dropping is implemented by interpolating the saliency mask to the spatial resolution of each layer. The integration of patch dropping and feature dropping further reduces model GOPs (in a hardware-friendly way) 
without compromising accuracy.

\subsection{{\underline{Model} pruning for computation reduction}}

The SOTA tracking model is overwhelmingly large.
It has been long proven that model pruning is an efficient way to reduce model complexity.
For effective and efficient hardware implementation, \textit{structured pruning} is much more beneficial compared to \textit{unstructured pruning}, as the latter introduces additional overhead such as sparse representation and format conversion. Furthermore, unstructured pruning hinders full utilization of parallel architectures~\cite{anwar2017structured}.
Therefore, we advocate for and propose structured pruning to fully accommodate our hardware design. 

\begin{table}
\centering
\caption{\small{The performance of irregularly pruned model (QDtrack) vs. pruning ratio (0\%, 80\%, 85\%, 90\%).  Here the sparse kernel ratio is given by   the ratio of pruned $3 \times 3$ kernel weights over the total number of pruned weights.
}
} 
\vspace{-1mm}
\label{tab:OMP}
\resizebox{0.45\textwidth}{!}{
\begin{tabular}{c|c|c|c|c}
\hline
\hline
\multirow{2}{*}{\begin{tabular}[c]{@{}c@{}}Metrics\end{tabular}} 
& \textbf{Dense Model} & \multicolumn{3}{c}{\textbf{Global pruning}} \\ 
\cline{2-5}  & \begin{tabular}[c]{@{}c@{}}0\% \end{tabular}
 & \begin{tabular}[c]{@{}c@{}} 80\%  \end{tabular}
 & \begin{tabular}[c]{@{}c@{}} 85\%  \end{tabular}
 & \begin{tabular}[c]{@{}c@{}} 90\%  \end{tabular}
 \\ \hline
      IDF1 ($\uparrow$)   &  0.714 & 0.712 & 0.706 & 0.703 \\
       MOTA ($\uparrow$)   &  0.637 & 0.631 & 0.627 & 0.624 \\
          \begin{tabular}[c]{@{}c@{}} Sparse kernel   ratio \end{tabular}      & 0\% & 
          36.59\% & 34.44\% & 32.53\%
          \\
\hline
\hline
\end{tabular}}
\vspace*{-4mm}
\end{table}
\subsubsection{Kernel-wise sparsity embedded in irregular weight pruning.}
The SOTA pruning method is iterative magnitude pruning (IMP) \cite{han2015learning}, which enables aggressive weight sparsity   without loss of model accuracy, as justified by the lottery ticket hypothesis \cite{frankle2018lottery}.
In Table\,\ref{tab:OMP}, we apply IMP to the QDTrack model 
and indeed observe a competitive MOT accuracy, measured by IDF1 and MOTA, as illustrated in Sec.\,\ref{sec: exp}.  However,  the weight sparsity pattern achieved by IMP  is unstructured, leading to the compatibility issue with hardware.  
Spurred by this fact, we examine if  there exists a structured sparsity pattern hidden in the irregular weight 
sparsity achieved by IMP. 
As shown in  Table\,\ref{tab:OMP}, the sparse kernel ratio, \textit{i.e.},   the  ratio of pruned $3\times3$ kernel weights over the total number of pruned weights takes over 30\%  in different weight pruning ratios (from $80\%$ to $90\%$). 
This implies that kernel-wise sparsity is embedded   in irregular weight pruning on QDTrack.

\subsubsection{Integration with irregular pattern-aware pruning.} 
We leverage the irregular pattern-aware pruning method \cite{ma2020image} to enforce sparse weights that cannot be covered using kernel-wise sparse patterns. 
The rationale behind such irregular pattern-aware pruning
\begin{wrapfigure}{r}{35mm}
\centerline{
\includegraphics[width=.22\textwidth,height=!]{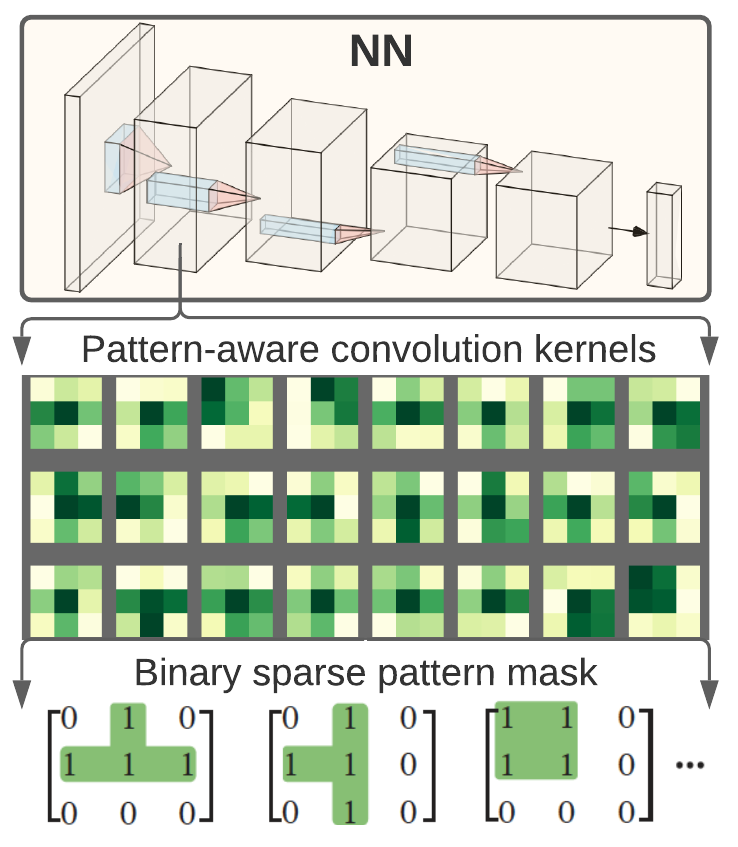}
}
\caption{\footnotesize{
Pattern  pruning. }}
\label{fig: pattern}
\vspace*{-3mm}
\end{wrapfigure}
is that  the effective area of a convolution kernel often maintains specific sparse patterns even if it does not yield kernel-wise sparsity, \textit{i.e.}, kernels with all zero weights. Thus, we follow \cite{ma2020image} to pre-define irregular sparse patterns \cite{niu2020patdnn} for $3 \times 3$ kernels, and leverage them to conduct  irregular but pattern-aware weight pruning. Note that pruning using a fixed number of pre-defined sparse  patterns facilitates  efficient hardware implementation compared with the conventional irregular weight pruning. 

Together with kernel-wise pruning, we propose the following hardware-aware pruning method:
    
$\bullet$ Call IMP to achieve the irregular weight sparse mask $\mathbf M_{\mathrm{IMP}}$;
   
$\bullet$    Extract kernel-wise sparsity mask  $\mathbf M$ from  $\mathbf M_{\mathrm{IMP}}$;
   
$\bullet$    Conduct irregular pattern-aware pruning on remaining weights identified by $(\mathbf 1-\mathbf M)$, then update $\mathbf M$ to obtain a new mask $\mathbf M^\prime$;
  
$\bullet$    Retrain non-zero model weighs under fixed pruning mask $\mathbf M^\prime$.

\subsection{\underline{Hardware} implementation and acceleration}
\label{sec:hw-acc}

\subsubsection{Patch dropping compatible tile-based accelerator design.}
Given HD input frames and the large ML model, the intermediate feature map for even a single layer is too large to be fully stored in the device's on-chip memory. Therefore, it is required that the input images and feature maps be partitioned into \textit{tiles} and load/compute tile-by-tile, as shown in Fig.~\ref{fig:hw-parallel}(a).
Fortunately, incorporating our patch dropping strategy, a tile-based accelerator design is beneficial since it allows the accelerator to completely skip the loading and computing of a tile if it is dropped at an early stage; this not only reduces the computation latency but also significantly reduces energy consumption, since off-chip data movement is much more power consuming than on-chip computation~\cite{chen2016eyeriss}.

\subsubsection{Pruning-aware and multi-level parallelism.}
\begin{figure}[b]
    \centering
    \includegraphics[width=.4\textwidth,height=!]{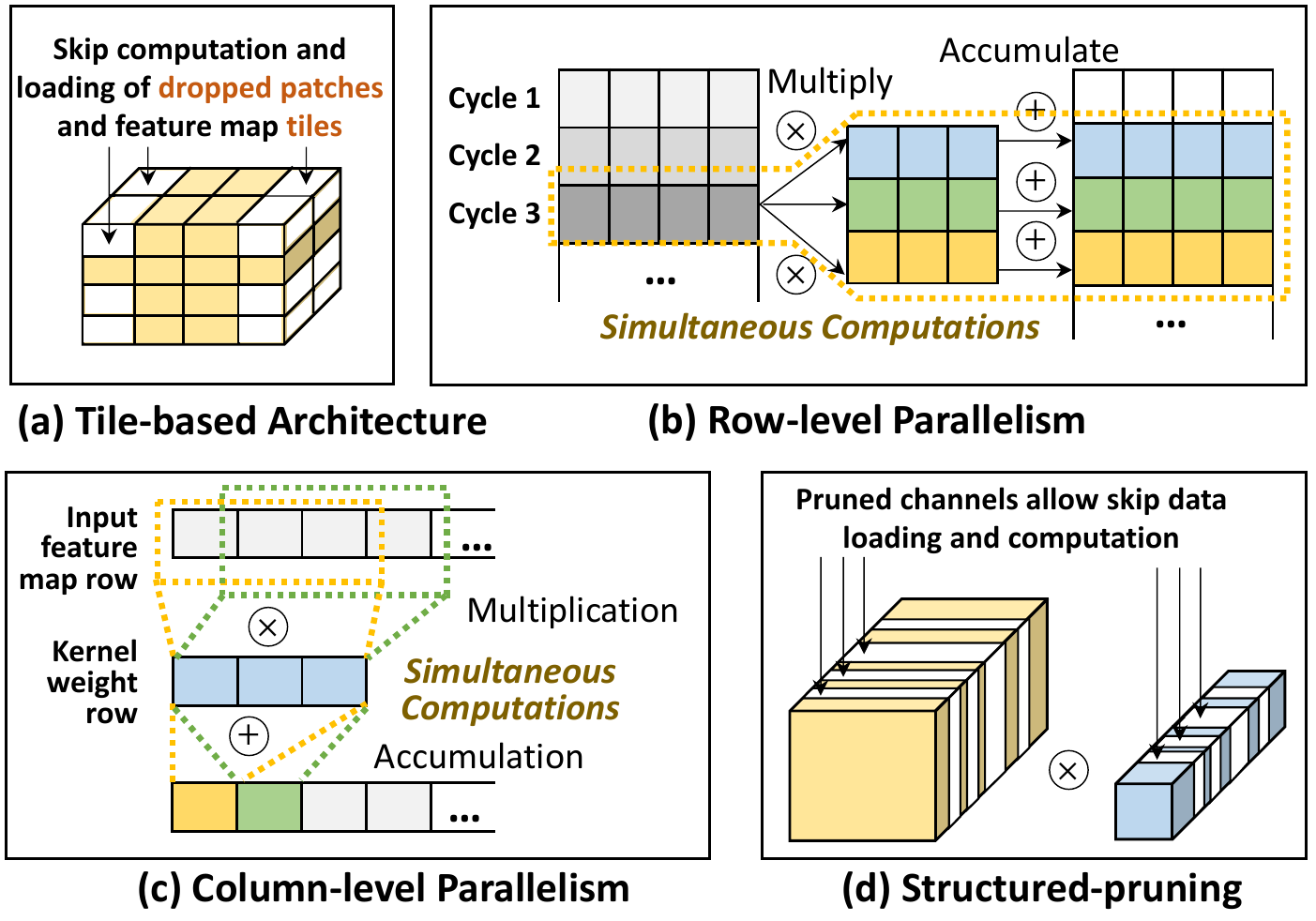}
    \vspace*{-3mm}
    \caption{\small{Pruning-aware parallelized hardware implementation.}}
     \label{fig:hw-parallel}
     \vspace*{-5mm}
\end{figure}

To achieve extremely high parallelism and to simultaneously exploit structured pruning, we propose a novel multi-level pruning-aware parallelization architecture, as illustrated in Fig.~\ref{fig:hw-parallel}. 
Consider one convolution layer as an example, where the input is a data tile of size $T_H\times T_W\times T_C$ along with a weight kernel of size $K\times K \times T_C$.
The commonly adopted parallel computation is along the $T_C$ dimension, i.e., $T_C$ channels are computed simultaneously ~\cite{hao2019codesign}.
However, such a parallelization scheme fails to employ the structured and pattern-based pruning, since a pruned-away channel in a kernel still needs to be computed with other channels in parallel. 
Therefore, in this work, we explore the parallelism along $T_H$ and $T_W$ dimensions in two levels, and exploit the sparsity along $T_C$ dimension.

\underline{First}, row-level parallelism, as shown in Fig.~\ref{fig:hw-parallel}(b), allows that at each clock cycle, one input row is computed with $K$ rows of a kernel, followed by self-accumulation for partial sum.
\underline{Second}, for column-level parallelism as shown in Fig.~\ref{fig:hw-parallel}(c), within the same clock cycle, $T_W$ multiplications are executed, followed by a $K$-to-$1$ adder tree.
\underline{Third}, since the row-level and column-level approaches have provided sufficient parallelism, we perform the convolution channel-by-channel sequentially without sacrificing computation latency. As shown in Fig.~\ref{fig:hw-parallel}(d), if one channel of a filter is pruned, 
we skip the data loading and computation for that channel, reducing energy and latency almost linearly. Specifically, with structured pruning, if $p\%$ channels are pruned away within a filter, the latency linearly reduces by $p\%$. Similarly, for pattern-based pruning, our proposed design 
uses a fixed sparsity pattern for the entire channel.

\subsubsection{Dataflow architecture using multi-FPGA for scalability.}
\begin{figure}[t]
    \centering
    \includegraphics[width=0.48\textwidth]{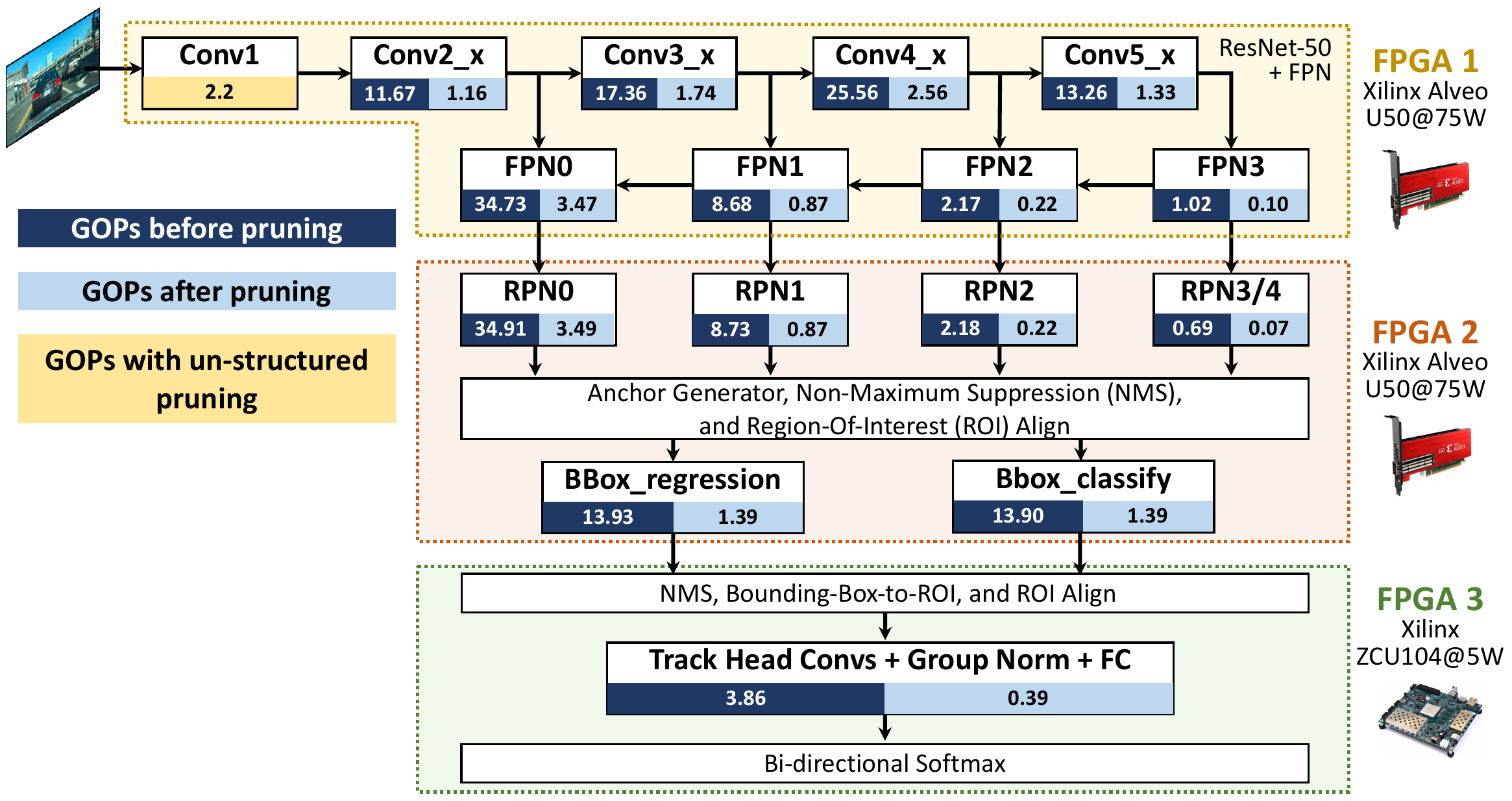}
    \vspace*{-3mm}
    \caption{\footnotesize{FPGA cluster for QDTrack implementation. FPGA 1 (Xilinx Alveo U50) is responsible for ResNet50 backbone; FPGA 2 (U50) is responsible for FPN and RPN modules; FPGA 3 (low-end ZCU104) processes the object tracking head. Under each module, we annotate the GOPs before/after pruning.}}
    \label{fig:FPGA-alloc}
\vspace*{-5mm}
\end{figure}

To overcome the device and design tool scalability issue, as mentioned in Sec.~\ref{sec:motivation-hw}, we propose a novel \textit{dataflow architecture} using multiple FPGAs, as illustrated in Fig.~\ref{fig:FPGA-alloc}.
Instead of executing the entire model on a giant FPGA with extremely high parallelism, which is not feasible under HD frames (limited by development toolkits), we dissemble the ML model into components and map them to different FPGAs. As demonstrated in Fig.~\ref{fig:FPGA-alloc}, multiple FPGAs work in a \textit{dataflow} manner: 1) The first FPGA executes the backbone ResNet-50 computation for one frame and meanwhile passes the intermediate feature maps to the second FPGA. Once the first FPGA finishes the backbone computation for one frame, it would start the computation for the next one. 2) The second FPGA starts the computation as soon as it receives the first tile of intermediate feature maps.  Such a dataflow architecture not only significantly reduces end-to-end latency but also increases the throughput by overlapping the computations of multiple frames. In addition, it resolves the scalability issue of both the device and the development tool as each FPGA can be developed independently.

\section{Experiments}
\label{sec: exp}

\subsection{{Experiment setup}}\label{sec: setup_exp}
We conduct the MOT task under the BDD100K dataset \cite{yu2020bdd100k},
one of the most representative and challenging self-driving car datasets that incorporates geographic, environmental, weather diversity and intentional occlusions.
It contains 100K videos with 30 frames per second. 
In our experiments, the training set has 1,400 videos and the validation set has 200 videos.

To evaluate the accuracy performance of our proposed tri-design approach, we adopt the standard MOT metrics,  ID F1 Score (IDF1)
and Multi-Object Tracking Accuracy (MOTA)
. These are two aggregated metrics widely used to evaluate the overall tracking performance by taking into account object detection accuracy and tracking consistency. In addition, we use the number of Identity Switches (IDSw) 
metric to evaluate the RL reward function in the first stage of data complexity reduction. 

To evaluate the efficiency of hardware implementation of our proposed tri-design approach, we demonstrate the QDTrack model on a dataflow cluster composed of three FPGAs: two mid-end Xilinx Alveo U50 boards (5,952 DSPs, 28 MB on-chip memory), and one low-end Xilinx ZCU104 (1,728 DSPs, 1.37 MB on-chip memory) board. The accelerator is developed using Vitis HLS 2021.1 and deployed using Vivado 2021.1. 

\subsection{Experiment results}

\begin{table*}[]
\vspace*{-4mm}
\begin{center}
\caption{\footnotesize{Performance comparison of our proposal with its variants and the   QDTrack \cite{pang2021quasi} baselines (implemented by GPU and FPGA respectively)   under various metrics including accuracy (IDF1 and MOTA) and hardware metrics given by on-board latency in the unit of millisecond, effective frame rate (EFR) in the unit of FPS, and power in the unit of Watt.
The data reduction is implemented using 
40\% temporal frame dropping together with 20\% spatial patch dropping.
The `$\times$' indicates the case of no data/model compression, and `{\NA}' means not applicable. 
The improvement of our proposed tri-design approach over GPU and FPGA baselines  are summarized.
} 
} 
\vspace{-3.5mm}
\label{table: overall}
\begin{threeparttable}
\small
\begin{tabular}{l|c|c|c|c|c|c|c|c}
\hline
\hline
\multirow{2}{*}{\begin{tabular}[c]{@{}c@{}}\textbf{Methods}\end{tabular}} & \multicolumn{2}{c|}{\textbf{Data/model compression}} & \multicolumn{5}{c}{\textbf{Metrics}} \\ 
\cline{2-9} 
& \begin{tabular}[c]{@{}c@{}} Data reduction \end{tabular}
& \begin{tabular}[c]{@{}c@{}}   Pruning \end{tabular}
& \begin{tabular}[c]{@{}c@{}} IDF1   ($\uparrow$) \end{tabular}
& \begin{tabular}[c]{@{}c@{}} MOTA  ($\uparrow$) \end{tabular} 
& \begin{tabular}[c]{@{}c@{}}   Latency  ($\downarrow$)   \end{tabular}
& \begin{tabular}[c]{@{}c@{}} EFR   ($\uparrow$)   \end{tabular}
& \begin{tabular}[c]{@{}c@{}} Power   ($\downarrow$)  
\end{tabular}
& \begin{tabular}[c]{@{}c@{}} Energy Efficiency   ($\downarrow$)  
\end{tabular}
 \\ \hline
 \textbf{QDTrack} (GPU baseline) 
 & $\times$ & $\times$ & 0.714  & 0.637 & 60.9  & 22.5 & 296 W & 13.2 J/frame \\
    \textbf{QDTrack on FPGA}  
    & $\times$ & $\times$ & 0.714  & 0.637 & 554.7  & 1.8 & 50.8 W & 28.2 J/frame \\
      \textbf{Variant}: Frame + patch drop
      & (40\%, 20\%) 
      &  $\times$ & 0.71   & 0.628  & 443.8   & 2.3 & 50.8 W & 22.0 J/frame \\
\rowcolor{Gray}
  \textbf{Tri-design} (ours)   & (40\%, 20\%)  & 90\% 
  & 0.704   & 0.617
  & 44.4   & 37.6  & 50.8 W & 1.35 J/frame \\ \hline
  \rowcolor{Gray}  \textbf{Improv. over GPU baseline}    &  \NA &  \NA  & -1.40\%   & -3.14\%  & \textbf{1.37$\times$}  & \textbf{1.67$\times$} & \textbf{5.83$\times$}& \textbf{9.78$\times$}\\
  \rowcolor{Gray}  \textbf{Improv. over FPGA baseline}    &  \NA &  \NA  & -1.40\%   & -3.14\%  & \textbf{12.5$\times$}  & \textbf{20.9$\times$} & \NA & \textbf{20.9$\times$} \\
\hline
\hline
\end{tabular}
\end{threeparttable}
\end{center}
\end{table*}

\subsubsection{{Results on video frame dropping.}}

We peer into the effect of the RL-based 
frame dropping model on our proposed tri-design pipeline.
We compare our frame dropping approach with two baselines: uniform dropping and random 
dropping. The former one drops 
the frames at a fixed time window for each  video, 
and the latter one randomly selects and drops frames in a video stream.
Fig.\,\ref{fig: compare_combine} shows the frame dropping performance at different drop rates from $15\%$ to $60\%$. As can be seen, our RL-based model generally 
outperforms the random drop model at different dropping ratios, indicating the 
effectiveness of the RL-based frame dropping model. We also observe that the `sweet' dropping ratio  is   achieved round $40\%$, which strikes the most graceful balance between spatial reduction and IDSw. In what follows, unless specified otherwise, we will assume that  $40\%$ temporal frame reduction is applied to MOT.

\subsubsection{{Results on saliency-based image patch dropping.}} 
\begin{figure}[t]
    \centering
    \begin{tabular}{cc}
     \includegraphics[width=.23\textwidth,height=!]{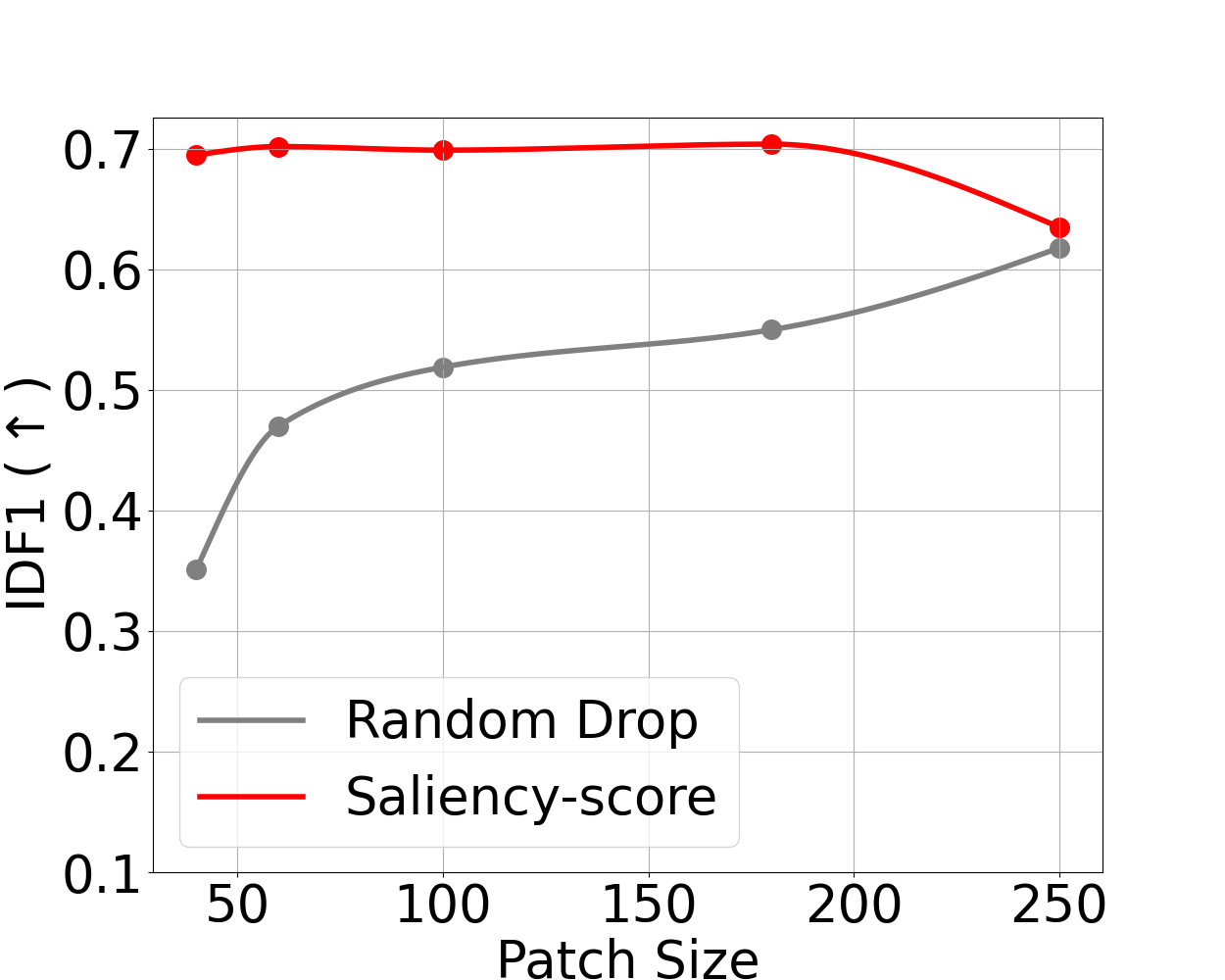}
     &
         \includegraphics[width=.23\textwidth,height=!]{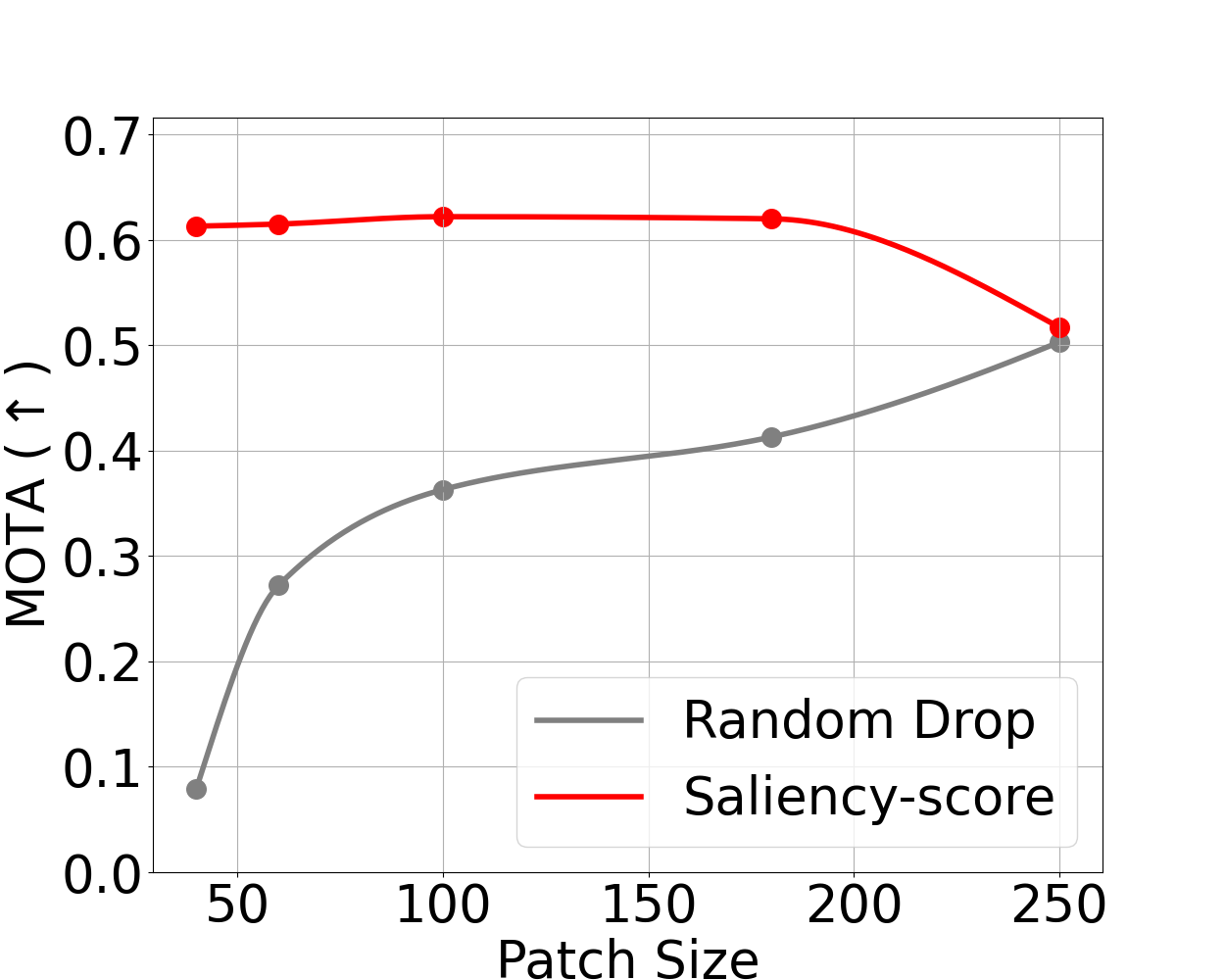}
    \end{tabular}
    \vspace*{-2mm}
    \caption{\footnotesize{Comparison between random patch dropping and our proposed saliency-based dropping vs. different drop patch sizes.} 
    }
     \label{fig:patchdrop}
     \vspace*{-4mm}
\end{figure}

We also studied the effect of spatial patch dropping in our framework. In Fig.\,\ref{fig:patchdrop}, we compare the saliency-based method with random dropping versus different patch drop sizes.
As can be observed, the proposed saliency-score based method consistently outperforms random dropping baseline at different patch size. We also observe that IDF1 and MOTA have similar performance for sizes smaller than 175 $\times$ 175, and they start to decrease if the drop patch size becomes larger than 175 $\times$ 175.
Therefore, we set the drop patch size as 60 $\times$ 60 in other experiments along with drop ratio as 20\% so as to maintain the IDF1 accuracy around 0.7.

\subsubsection{{Results on model pruning.}}
\begin{figure}
\centerline{
\includegraphics[width=.49\textwidth,height=!]{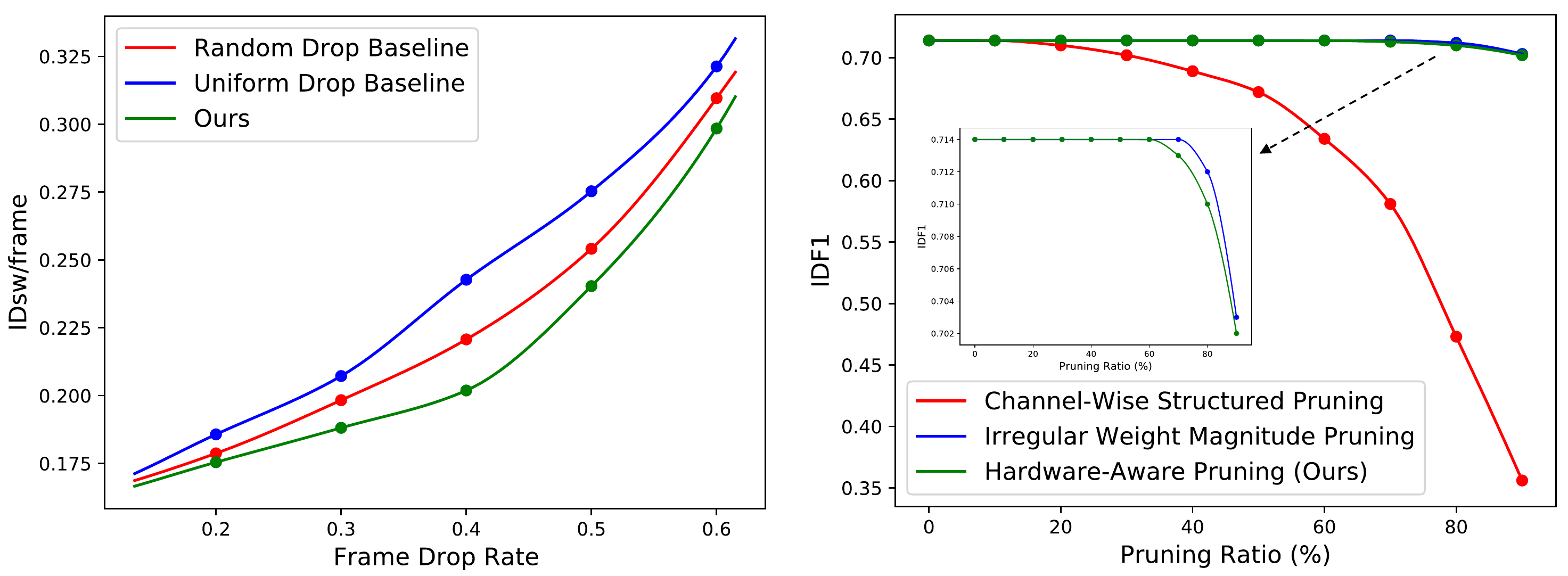}
}
\vspace*{-3mm}
\caption{\footnotesize{
Performance comparison: \textbf{Left:} Temporal filtering; \textbf{Right:} Pruning. 
}}
\label{fig: compare_combine}
\vspace*{-8mm}
\end{figure}

We compare our proposed model compression method with two commonly-used pruning baselines: irregular weight magnitude pruning (IMP) and channel-wise structured pruning. The results are demonstrated in Fig.\,\ref{fig: compare_combine}. 
As we can see, channel-wise pruning is quite sensitive to the pruning ratio. By contrast, our proposed hardware-aware pruning method  is capable of achieving  MOT performance similar to IMP across different pruning ratios. 
Additionally, in Table \ref{table: overall} we show that when integrating  with the spatiotemporal data reduction algorithms, 
our proposed hardware-aware pruning method yields back-end acceleration according to the performance of on-board  latency and the effective video frame rate.

\subsubsection{{Tri-design performance.}}

We next summarize the performance of integrated frame dropping, patch dropping, and model compression.
In Table\,\ref{table: overall}, we demonstrate both MOT accuracy  and hardware implementation efficiency metrics of our proposed tri-design approach.
The baselines include the original QDTrack~\cite{pang2021quasi} implementation on GPU Tesla V100 and the Xilinx FPGA Alveo U50 cluster (Section~\ref{sec:hw-acc}, Fig.~\ref{fig:FPGA-alloc}).
We present the results of different configurations of data complexity reduction and model weight pruning; each configuration results in one variant of the proposed tri-design method.
\underline{Accuracy}. As we can see,  
the use of proposed data complexity reduction strategy (\textit{i.e.}, the row `Variant')
does not  hamper the  MOT accuracies in terms of IDF1 and MOTA scores. 
\underline{Throughput}.
Comparing to GPU baseline, 
our tri-design approach reduces latency by $1.37\times$ and improves the effective frame rate (EFR) by 1.67$\times$;
comparing to FPGA baseline,
our tri-design together with proposed sparsity-aware multi-level parallelism reduces latency by $12.5\times$ and improves the EFR by $20.9\times$, demonstrating significant performance boost.
\underline{Energy Efficiency}.
Meanwhile, our approach shows remarkable energy efficiency.
The baseline GPU Tesla V100 has a power consumption of 296W, while our FPGA cluster has only 50.8 W, leading to $5.83\times$ power reduction. Together with 1.67$\times$ higher frame rate, the energy efficiency is 9.78$\times$ higher ($5.83\times 1.67$). To summarize, our approach outperforms GPU and FPGA baselines in all hardware efficiency metrics with negligible accuracy loss.

\section{Conclusions}
We introduce \textit{data-model-hardware tri-design} for MOT implementation on the edge devices, which exploits aggressive data reduction, model compression, and ultra-low-power hardware innovation for a hardware-aware  ultra-light   algorithm development. We demonstrate the effectiveness of the proposed tri-design through extensive experiments.
Compared to SOTA MOT baseline, our approach on Alveo U50 FPGAs achieves 12.5$\times$ latency reduction, 20.9$\times$ effective frame rate improvement, $5.83\times$ lower power, and $9.78\times$ better energy efficiency comparing to Tesla V100 GPU.

\bibliographystyle{ACM-Reference-Format}
\bibliography{tri}

\end{document}